\documentclass[conference]{IEEEtran}

\usepackage{graphicx}

\ifCLASSINFOpdf
 
\else
 
\fi

\begin{document}

\title{Analyzing Gender Polarity in Short Social Media Texts with BERT: The
Role of Emojis and Emoticons}

\author{\IEEEauthorblockN{Saba Yousefian Jazi}
\IEEEauthorblockA{Departmet of Computer Science and \\ Engineering\\
University of North Texas\\
Denton, Texas 76205\\
Email: sabayousefianjazi@my.unt.edu}
\and
\IEEEauthorblockN{Amir Mirzaeinia}
\IEEEauthorblockA{Departmet of Computer Science and \\ Engineering\\
University of North Texas\\
Denton, Texas 76205\\
Email: amir.mirzaeinia@unt.edu}
\and
\IEEEauthorblockN{Sina Yousefian Jazi}
\IEEEauthorblockA{Independent Researcher \\
Santa Clara, California 95051\\
Email: sinayousefian@gmail.com}}

\maketitle

\begin{abstract}
In this effort we fine tuned different models based on BERT to detect the gender polarity of twitter accounts. We specially focused on analysing the effect of using emojis and emoticons in performance of our model in classifying task. We were able to demonstrate that the use of these none word inputs alongside the mention of other accounts in a short text format like tweet has an impact in detecting the account holder's gender.
\end{abstract}

\IEEEpeerreviewmaketitle

\section{Introduction}
One of the public and big size data source for textual data is tweeter. This platform initially was designed to gives users the opportunity to send a short textual input in the platform. Although the platform has evolved into much more input types, and increased the limit of tweets to 280 character, the main functionality for the users still stays as short texts.Along with the popularity of this platform, the developed API's provides the opportunity to scrap and analyze this textual data. Even though after February 2023, access to such data is not free, there are some available dataset collected from tweeter that still could be useful.
One of the main concerns that comes with every public social media platform, is the security and privacy of such platforms. In 2020, around 130 high profile tweeter accounts were hacked. While some were immediately disabled, others were used to tweet unauthorized notes. One of the motivations of this effort is to help the security teams in detecting such incidents. If a model can successfully detects that the input text does not match the authors style, it can flag such activity and proceed further actions. One of the main indications and simplest step in such process, could be detecting if the gender of author matches the users gender.  Aside from the security purposes, there are interests in marketing and financial sectors regarding social analysis of social media users, including tweeter. Detecting a simple but private information such as gender of users from texts may seem trivial at first, but most of the developments in Large language models are focused on large texts. In contrast to large texts with high word count, short texts may not provide as much context for pretrained models such as BERT. That is why we hypothesize that in such scenarios we need to use style parameters as much as possible.   
 Specifically, inn this paper we investigate the answer to this question: given a short text document, can we identify if the author is a man or a woman based on their style of tweeting?

\section{Background}
Not much research delves into analyzing the diverse styles found in short texts from different users. Some researchers have tackled similar tasks, like digging into profiles, as seen in studies  \cite{nia2022twitterbased}, \cite{Male2021}, and \cite{Vikas2022}. They sifted through thousands of profiles, studying how people described themselves using various subjects, word choices, and qualifications. While these studies provided valuable insights into a wide range of users, including non-binary ones, they often overlooked the nuances of text styles. Profiles, being usually short and lacking full sentences, sometimes relied on profile pictures for predictions.

One of the closest attempts at detecting gender from tweets was done in \cite{9180161}. They tried different methods, like bag-of-words and traditional machine learning, to understand short texts better. But they missed out on emojis and emoticons, which we think could be crucial in understanding how users communicate. Another attempt in Twitter author profiling was \cite{ouni2022bots}, one of the few to consider both text styles and elements like mentions and emoticons, though their main focus was on telling apart robot and non-robot profiles and they used different methods. In 2023, researchers in \cite{onikoyi2023gender} published another approach to incorporate a user’s profile description on Twitter to their tweets to identify the gender. However, they applied more classical machine learning methods in their experiments. These days, more recent transformer-based models (such as BERT. \cite{devlin2018bert}) have been shown   to be more effective than classical approaches in these applications. We mostly focused on how parameters tied to style, such as mentions and emojis  can improve pretrained models such as BERT to quickly adapt to new tasks on short texts.

As social media grows, so does the issue of virtual aggression. There have been limited efforts to detect gender based on profanity or hate speech in tweets, like the works shown in \cite{wong2020different} and  \cite{ijerph18084055}. While these studies mostly aimed to find out which types of users engaged in negative communication, they also tried to define and analyze their target groups. However, like previous efforts, they often simplified initial data, leaving out details such as emoticons, connections, and mentions among different users.

\begin{table*}
\caption{Distribution of career categories in dataset}
\centering
\begin{tabular}{lll}
\hline
\textbf{Career} & \textbf{Female} & \textbf{Male}\\
\hline
\verb|Singer |& \verb|13| & \verb|9| \\
\verb|Actor/Actress |& \verb|5| & \verb|1| \\
\verb|Media personality |& \verb|3| & \verb|4| \\
\verb|Athlete/sport analyst |& \verb|0| & \verb|2| \\

\hline
\end{tabular}

\label{tab:Singer} 
\end{table*}
\section{Pretrained Model}
For the purpose of this effort, we used a Bidirectional Encoder Representations from Transformers (BERT) architecture. BERT is one of the advanced models that is developed for large language model applications published in 2018 \cite{devlin2018bert}. 
The BERT model excels in tasks such as text classification, question answering, sentiment analysis, and language translation. BERT is pre-trained on vast amounts of text data using unsupervised learning techniques that helps BERT to learn rich and comprehensive representations of words and sentences, enabling it to understand language nuances, semantics, and relationships effectively.
The model we selected is a BERT Base uncased model, pre-trained with 110 million parameters for different tasks including classification. We also used the BERT tokenizer and BERT embedding to feed our input into the model. We then added a drop out layer with the rate of 0.1\% as drop out percentage. We then added the dense layer with the sigmoid function to perform the binary classification on our dataset. for the training process, we achieved the best results with learning rate of 2e-5. In each experiment, we trained the classifier for 10 epochs on Tesla T4 GPU with each of them taking 25 minutes to run on our dataset.
\section{Experiments}
Although we used the BERT preprocessing layer for our input, we did some preprocessing before feeding the data to the first layer.
We first removed the retweets from the tweet dataset. These samples are easily recognized since they start with "b\textbackslash RT". Removing these tweets decreased the number of tweet samples significantly. We also shuffled the dataset to make sure that before sampling dataset for validation portion, samples are mixed properly. We selected 25\% of our dataset for validation. We checked both training and validation set to have similar distribution as original dataset, regarding the male and female percentage, mean tweet length and max tweet length. After the initial cleaning, we applied our functions to turn emojis and emoticons into representing texts. Since the data is collected and read as UTF-8, the emojis will follow this pattern: a 4 character- sub-string starting with backslash and a lower case alphabet, then an upper case alphabet and then 1 digit. Table \ref{tab:Emojis}  showcases a few samples of these emojis. In total 842 emoji codes were recognizable and we ran the function to replace them with their equivalent text that defines the emoji. 

\subsection{Data}
Before february 2023, there were multiple ways one could use to scrap twitter data. The first approach could be using the available python libraries such as Tweepy to access the standard Twitter API. Although the Twitter API was easily accessible via a simple registration, there were still some limitation regarding scrapping the Twitter data. The first limitation, comes to the timeline of tweets. Using this method, one could only scrap the tweets up to 7 days ago. This limitation made it very hard to gather a large deadset in a short amount of time. Another limitation of such library is that one can only scrap the last 3200 of an account.  After 2023, gathering data was made even more cumbersome, as one needs to pay for such access added to the other limitations. Considering these limitations and many more, we decided to go with one of the big twitter detests that is gathered and shared with public. This dataset is called " 1000 celebrity tweets" and is available in kaggle dataset section. This dataset that is updated monthly, consists of tweets from the top 1000 most followed celebrity accounts in twitter. The number of tweets for each account is very different and can goes from twenty to four thousands tweets per an account. 
Also the timeline of tweets varies depending on how frequent the account holders tweet and how old their account is. Another important attribution about this dataset is that it does not separate English or non-English speaker and the dataset contains many different language speaker accounts.

For our project, we separated 21 female English speaker accounts and 16 male English speaker accounts. The gender of these accounts were confirmed from the biography of them or Wikipedia information. The reason for different number of male and female account holders is that we wanted to make sure that the distribution of tweets in the combined dataset will remain as 50\% for each gender. As the career distribution of our dataset accounts is shown in Table \ref{tab:Singer}, we can see that most of these accounts belong to singers and this type of accounts disproportionately effects the type of tweets that we have at our disposal.
\begin{figure}
    \includegraphics[width=8cm]{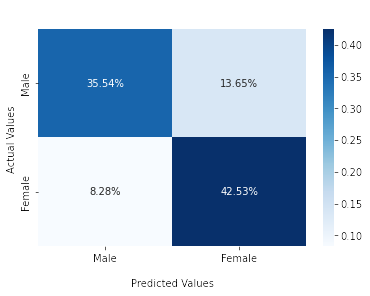}
    \caption{ Confusion matrix of original experiment: replacing the emojis with text and including mentions. }
\end{figure}
\subsection{Experiment with mentions}
After analysing the results and some of the wrong labeled samples, we noticed that the presence of mentioning of other accounts, which  starts with "$@$", be it a person or not, would cause the model to change its prediction. To test our hypothesis, we altered the same training dataset to be free of mentions and run the same process of fine tuning and validation. As it is shown in table \ref{tab:results}, we can see that our hypothesis regarding presence of mentions effecting the performance of model was correct.
\subsection{Sentiment analyses}
After running the model on validation set, we analysed some of the mistakes and correct labeled samples. We noticed that there is a small difference between the male tweets wrongfully labeled from the female tweets wrongfully labeled. We noticed that the female tweets that have been wrongfully labeled as male, had relatively shorter distribution than the correct labeled ones. It is not worthy to mention that we also noticed that male tweets that were labeled correctly as male had also shorter length.

\begin{table}
\caption{Emojis and their UTF-8 codes in tweet samples.}

\centering
\begin{tabular}{cc}
\hline
\textbf{code} & \textbf{text replacement}\\
\hline
\verb|\xF0\x9F\x98\xA0| & {ANGRY FACE} \\
\verb|\xF0\x9F\x98\x89| & {WINKING FACE} \\
\verb|\xF0\x9F\x98\xA2| & {CRYING FACE} \\
\verb|\xF0\x9F\x98\xAB| & {TIRED FACE} \\
\verb|\xF0\x9F\x98\xB5| & {DIZZY FACE} \\

\hline
\end{tabular}
\begin{tabular}{lc}
\hline
\hline
\end{tabular}
\label{tab:Emojis}
\end{table}
Since most of the emojis and emoticons are replaced with words with high intensity in emotions, we decided to analyse our validation samples from the sentiment view. To achieve that goal, we used “text2emotion” library to analyse the test samples to see if there is a correlation between the emotions perceived from tweet after replacing the emojis and the prediction or not. This library is able to categorize text into 5 classes of: Angry, Fear, Happy, Sad and Surprise.
The results showed that the tweets that were not categorized as any emotion, or neutral emotion were mostly labeled as male and tweets that were categorized as fear or happy were mostly categorized as female.
\begin{figure}
    \includegraphics[width=8cm]{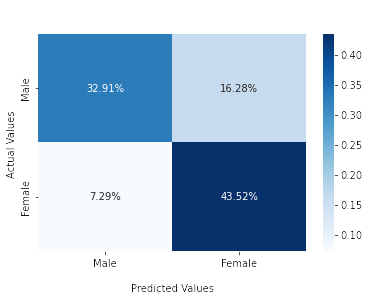}
    \caption{ Confusion matrix of second experiment: replacing the emojis with text and removing mentions. }
\end{figure}
\section{Discussion of results}
Although we were able to replace more than 800 emojis, there were still some emojis that were not replaced with text. That is because new emojis are developed and added to the collection of available emojis everyday and it is hard to keep track of every new developed emojis and their code and textual meaning. We also think that the text that is used as replacement may have an effect in giving the whole sample a positive or negative sentiment, even though the intensity of the original text without emoji is not comparable with it.
Another challenging factor that we have to keep in mind is that since these celebrities mostly have a person or a team responsible for public relations, there may be a good chance that many of their tweets are not entered by themselves or being checked, controlled or filtered before sending to the public. Because of that, the personality of a person may not fully be portrayed in their tweets. Also, a good portion of tweets from the TV celebrities, are not specifically representative of the celebrity in that show and mostly about the show. That also happens when the account belongs to a band or any team effort. Similar Case happens for singers or actors that tweet about their new album or movie. Although they may be tweeting themselves, but since these tweets mostly have an advertisement schema, they do not provide any useful information and that could be some sort of redundant data.
\section{Conclusion}
\begin{table}
\caption{Comparison of results in different model}

\centering
\begin{tabular}{lccc}
\hline
\textbf{code} & \textbf{Accuracy} & \textbf{Precision} & \textbf{Recall}\\
\hline
\verb|With mention+no emoji| & {0.7764} & {0.7841} & {0.7728} \\
\verb|With mention+with emoji| & {0.7807} & {0.7571} & {0.8370}\\
\verb|No mention+no emoji| & {0.7813} & {0.7600} & {0.7904}\\
\verb|No mention+with emoji| & {0.7901} & {0.7663} & {0.8001} \\
\hline
\end{tabular}
\label{tab:results}
\end{table}

Tweeter is one of the semi-free available textual sources that could be used for many language analyses and NLP modelings. We believe that there is a lot to be done regarding processing the short text. Although recent developments in attention based models such as BERT has hugely improved the results in many tasks, these models are mostly not as great when it comes to shorter texts. With all that being said, we believe that there is a need to analyse how big is the impact of replacement text in sentiment of texts. There is also a need to gather the samples from accounts that may tweet more freely and are able to express their thoughts and feelings more uncensored. Also, it is better to focus on diversifying the authors type of activities or career since that has a huge impact on what they tweet about too.

\bibliographystyle{ieeetr}
\bibliography{references}

\end{document}